\documentclass{article}
\usepackage{ijcai25}

\usepackage{times}
\usepackage{soul}
\usepackage{url}
\usepackage[hidelinks]{hyperref}
\usepackage[utf8]{inputenc}
\usepackage[small]{caption}
\usepackage{graphicx}
\usepackage{amsmath}
\usepackage{amsthm}
\usepackage{booktabs}
\usepackage{algorithm}
\usepackage{algorithmic}
\usepackage{enumerate}
\usepackage{amsfonts}
\usepackage[switch]{lineno}
\usepackage[textsize=tiny]{todonotes}
\usepackage{xspace}
\usepackage{multirow}
\usepackage{enumerate}

\def\HH{{\mathcal H}}

\def\OO{{\mathcal O}}
\def\PP{{\mathcal P}}

\def\XX{{\mathcal X}}

\newcommand{\bI}{\mathbb{I}}
\newcommand{\cH}{\mathcal{H}}

\newcommand{\cA}{\mathcal{A}}

\newcommand{\cG}{\mathcal{G}}
\newcommand{\cT}{\mathcal{T}}

\usepackage{mathtools}

\newcommand{\algo}{\textsc{Pmi}\xspace}

\def\Setup{{\textsc{CGi}}\xspace}

\title{\Setup: Identifying Conditional Generative Models with Example Images}

\author{
Zhi Zhou$^{1,2}$
\and
Hao-Zhe Tan$^{1,3}$\and
Peng-Xiao Song$^{1,4}$\And
Lan-Zhe Guo$^{1,3,}$\thanks{Corresponding author.}
\affiliations
$^1$National Key Laboratory for Novel Software Technology, Nanjing University, China\\
$^2$School of Computer Science, Nanjing University, China\\
$^3$School of Intelligence Science and Technology, Nanjing University, China\\
$^4$School of Artificial Intelligence, Nanjing University, China\\
\emails
\{zhouz, tanhz, songpx, guolz\}@lamda.nju.edu.cn}

\begin{document}

\maketitle

\begin{abstract}
    Generative models have achieved remarkable performance recently, and thus model hubs have emerged.
    Existing model hubs typically assume basic text matching is sufficient to search for models.
    However, in reality, due to different abstractions and the large number of models in model hubs, it is not easy for users to review model descriptions and example images, choosing which model best meets their needs.
    Therefore, it is necessary to describe model functionality wisely so that future users can efficiently search for the most suitable model for their needs.
    Efforts to address this issue remain limited. In this paper, we propose \emph{Conditional Generative Model Identification} (\Setup), which aims to provide an effective way to identify the most suitable model using user-provided example images rather than requiring users to manually review a large number of models with example images.
    To address this problem, we propose the \emph{Prompt-Based Model Identification} (\algo) , which can adequately describe model functionality and precisely match requirements with specifications. To evaluate \algo approach and promote related research, we provide a benchmark comprising 65 models and 9100 identification tasks. Extensive experimental and human evaluation results demonstrate that \algo is effective. For instance, 92\% of models are correctly identified with significantly better FID scores when four example images are provided.
\end{abstract}

\section{Introduction}

% 模型库变大，用户检索越发困难
Deep generative models~\cite{jebara2012machine}, such as variational autoencoders~(VAE)~\cite{VAE-KingmaW13,VAE-KingmaW19,VAE-ParmarLLT21}, generative adversarial networks~(GAN)~\cite{GAN-GoodfellowPMXWOCB14,GAN-SohnLY15,GAN-CreswellWDASB18}, flow-based models~\cite{FM-RezendeM15}, and diffusion models~\cite{SD-DicksteinW15,SD-DhariwalN21,SD-Rombach22}, have shown significant success in image generation. Generative model hubs like Hugging Face and CivitAI have been established to facilitate the sharing and downloading of models by developers and users, respectively. 
However, with the rapid growth of available models, finding the most suitable model for specific tasks has become a critical challenge.

% 现有检索技术太简单
Existing generative model hubs provide basic methods such as model tag filtering, text matching, and download volume ranking~\cite{GPT-abs-2303-17580} to help users search for models. However, the complex functionalities of generative models cannot be adequately described using only textual and statistical information~\cite{lv22cbs,luo2024stylus}. 
Although model hubs like CivitAI and OpenArt offer example images for each model to assist users in finding their desired models, this solution has two main limitations. 
First, example images cannot fully represent model functionality, especially when user purposes and offered examples mismatch.
Second, users still need to manually review example images repeatedly, which is time-consuming and heavily relies on their expertise.
Therefore, model selection remains challenging for users, as they must iteratively review, download, and test multiple models before finding a suitable one.

\begin{figure}[t]
  \centering
  \resizebox{\linewidth}{!}{
  \includegraphics{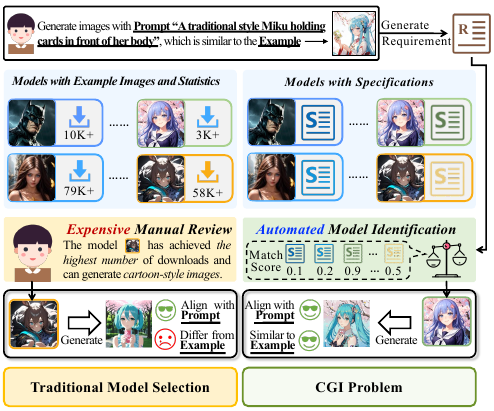}
  }
  \caption{
      Comparison between traditional model selection and \Setup problem setting. With the design of the specification and model identification, the most suitable model can be efficiently identified by matching the user's requirements with the model's specification.}
  \label{fig:setting}
  \vspace{-0.3cm}
\end{figure}

\textbf{\underline{C}onditional \underline{G}enerative Model \underline{I}dentification (CGI)}. 
The above limitation inspires us to consider the following question: 
\emph{Can we describe the functionality of conditional generative models in a precise format that enables efficient and accurate model identification by matching their functionalities with user requirements?}
Following the learnware paradigm~\cite{LW-Zhou22}, we call the functionality description the model's specification. 
An important characteristic of this problem is the requirement to assign a specification to the model upon uploading it to the model hub which allows future users to simply compute the similarity between the specification and their requirements to search for models. 
This is fundamentally different from the scenario where a query and a large set of candidate models are provided, and the objective is to learn the ranking. 
We call this novel setting \emph{\underline{C}onditional \underline{G}enerative Model \underline{I}dentification} (\Setup). To the best of our knowledge, this problem has not been studied yet. Figure~\ref{fig:setting} presents an illustration of the CGI problem and different from the traditional model selection process. 

\textbf{Challenges of CGI}. 
The challenges of CGI come from two main sources: (1) How to assign specifications to adequately describe the functionalities of different conditional generative models as well as the user requirements, and (2) How to match the users' requirements with the models' specifications in the same space. There are few works related to this problem. For instance, HuggingGPT~\cite{GPT-abs-2303-17580} and Stylus \cite{luo2024stylus} proposed to describe the model's functionality using natural language and search for models using ChatGPT. However, only natural language is insufficient to describe the specialty of the model since it is not easy to match the model's specific functionality with the text description. \cite{LW-Wu23} proposed to describe the functionality of discriminative models by approximating the training data distribution. However, for conditional generative models, the functionality is not only related to the generated data distribution but also to the prompts. There is no existing work that can be directly applied to solve CGI problems, requiring to study more wise descriptions for generative models.

\textbf{Our Solution}. To this end, we present a novel systematic solution, namely, \underline{P}rompt-Based \underline{M}odel \underline{I}ndetification (\algo). 
Specifically, we first introduce \emph{Automatic Specification Assignment} to generate specifications using a pre-defined prompt set or developer-provided prompt set. Then, \emph{Requirement Generation} abstracts the user requirements. 
Both specification and requirement are projected into a unified model matching space for future model identification. 
Finally, we propose a \emph{Task-Specific Matching} mechanism to adjust specification according to the user requirements in the model matching space to precisely identify the most suitable model. 
The basic idea is that if the model produces a data distribution that is very close to the user's example images with the similar prompts, then there is a high probability that the model is useful. 
To evaluate the effectiveness of \algo and promote the related research, we developed a benchmark comprising 65 conditional generative models and 9100 model identification tasks. Extensive experiment results demonstrate that \algo is effective. 
Moreover, human and GPT evaluation results confirm both the validity of our evaluation protocol and the superior performance of \algo. 
Our main contributions can be summarized as follows:
\begin{enumerate}[(a)]
    \item We introduce a novel setting called \Setup\ for identifying conditional generative models that match user requirements. This problem setting consists of two key challenges: (1) the construction of model specifications and user requirements and (2) the matching of user requirements to model specifications.
    \item We analyze the challenges inherent in the \Setup\ problem and propose an effective solution \algo. Our approach can adequately describe the functionality of generative models and enables future users to efficiently find the most suitable model with just a few example images. 
    \item We develop a benchmark with 65 models and 9100 identification tasks to evaluate model identification approaches. Extensive experiments and human evaluation results demonstrate that our proposal can achieve satisfactory model identification performance.
\end{enumerate}

\section{Related Work}

This study is related to the following two aspects:

\paragraph{Learnware and Model Selection.} With the development of model hubs, users have the option to search for and adapt pre-trained models that satisfy their specific needs. 
A similar model selection approach can also be applied to enhance generalization~\cite{zhou2024decoop,zhou2025fta}.
Learnware~\cite{Zhou16learnware} offers a paradigm to identify models for the users. 
Recently, Wu~\cite{LW-Wu23} proposed to describe the model's functionality by approximating the training data distribution and searching for models by comparing the approximated distribution distance. Guo~\cite{LW-Guo23} proposed to describe the model's functionality by approximating the model's parameters with a linear proxy model and enabling the model search by comparing the proxy model's parameters. However, these methods are not applicable to the generative model search. 
A few works are related to the generative model search. For example, \cite{GPT-abs-2303-17580} proposed to describe the model's functionality via natural language (e.g., model tags, model architectures, resources requirement) and adopted ChatGPT as a model selector to search useful models that meet user's requirements from the HuggingFace platform. 
Stylus~\cite{luo2024stylus} describes the model functionality with vision-language models, and identify conditional generative models with large language models. 
However, only natural language can not describe the model's functionality adequately, more precise description needs to be studied. 
\cite{lv22cbs} proposed a content-based search method that can be applied to unconditional generative models. 
Therefore, existing works are not applicable to the \Setup problem, and this paper presents the first attempt to solve this problem.

\paragraph{Generative Models.} In recent years, generative models have become one of the most widely discussed topics in the field of artificial intelligence for their promising results in image generation, exemplified by models such as Generative Adversarial Networks~\cite{GAN-GoodfellowPMXWOCB14,GAN-ArjovskyCB17,GAN-BrockDS19,GAN-ChoiUYH20}, Variational Autoencoders~\cite{VAE-KingmaW13,VAE-OordVK17,VAE-VahdatK20}, Diffusion Models~\cite{DM-NicholD21,SD-DhariwalN21,SD-Rombach22}, etc. With the development of the generative model, various generative model hubs, e.g., HuggingFace and Civitai, have been developed to enable model developers to share models. These numerous generative models show different specialties and functionality. Our goal is not to introduce a new model. Instead, we want to study a new mechanism that can well organize the developed models and enable future users to efficiently find the most suitable one.

\section{Preliminary}

In this section, we first introduce the problem setup of \Setup\ problem. Then, we present the problem analysis to show the core challenge of \Setup\ problem.

\subsection{Problem Setup}
Assume the model hub has $M$ conditional generative models $\left \{ f_m \right \}_{m=1}^M$. Each model is associated with a corresponding specification $S_m$ to describe its functionalities for future model identification. There are two stages in the CGI setting: the \emph{submitting stage} for model developers and the \emph{identification stage} for future users.

\paragraph{Submitting Stage.} The model developer submits a model $f_m$ to the model hub, and then we assign a specification $S_m$ to the model. 
Formally, the specification $S_m$ is generated by a specification assignment algorithm $\mathcal{A}_s$ using the model $f_m$, i.e., $S_m = \cA_s \left (f_m \right )$.
It is important to note that uploaded models are anonymous with no mandatory constraints, which means we cannot access their training data and developers are not guaranteed to provide required model information.

\paragraph{Identification Stage.} 
For any user task $\tau$, models are identified from the model hub using one or a few example images $X_{\tau}=\{x_i^{\tau}\}_{i=1}^{N_{\tau}}$. When users upload example images to describe their needs, the model hub generates the requirement represented in the specification space $R_{\tau} = \mathcal{A}_r(X_{\tau})$ using a requirement generation algorithm $\cA_r$. Then, we match the requirement $R_{\tau}$ with model specifications $\left \{ S_m \right \}_{m=1}^M$ using an evaluation algorithm $\cA_e$ and compute the matching score $\widehat{s}^{\tau}_m = \cA_e(S_m, R_{\tau})$ for each model $f_m$.
Finally, return the best-matched model with the maximum score or a list of models sorted by $\left \{ \widehat{s}^{\tau}_{m} \right \}_{m=1}^M$ in descending order.

The two main problems for addressing \Setup\ are:
(a) How to design $\cA_{s}$ and $\cA_{r}$ to fully characterize the functionality of submitted conditional generative models and user requirements? 
(b) How to design $\cA_{e}$ to effectively identify the most suitable model for users' specific needs using the specifications and requirements?

\subsection{Problem Analysis}

Learnware~\cite{Zhou16learnware} provides an effective framework for describing the functionality of discriminative models. 
It describes model functionality by approximating the training data distribution and performs model search by matching the approximated training and testing data distributions.

Specifically, the learnware methods~\cite{LW-Wu23} use Kernel Mean Embedding (KME) techniques to represent training data distributions by mapping a probability distribution $\mathbb{P}$ defined on $\XX$ into a reproducing kernel Hilbert space (RKHS) as
\begin{equation}
    u_k(\mathbb{P}) \coloneqq \int_{\XX}k(x, \cdot)d\mathbb{P}(x)
\end{equation}
where $k: \XX \times \XX \to \mathbb{R}$ is a kernel function with associated RKHS $\HH$.
In cases of finite training data, the empirical KME can be used to approximate the true KME using the dataset $X=\{x_i\}_{i=1}^{N}$ which can be seen as data points sampled from the distribution $\mathbb{P}$: 
\begin{equation}
    \hat{u}_k(\mathbb{P}) \coloneqq \frac{1}{N}\sum_{i=1}^{N}k(x_i, \cdot)
\end{equation}
It has been proved that the empirical KME can converge to the true KME at a rate $\OO(1/\sqrt{N})$~\cite{SmolaGSS07}.

However, existing learnware methods cannot be directly used for the \Setup\ problem since they were designed for discriminative models rather than conditional generative models. 
Corresponding to the two main problems mentioned above, the two main challenges in extending learnware methods to the \Setup\ problem are:
\begin{enumerate}[(a)]
    \item How to project both the complex functionality of conditional generative models and the diverse styles of user example images into a unified space for future model identification?
    \item How to design an effective matching mechanism between user requirements and model functionality corresponding to  specific user tasks?
\end{enumerate}

\begin{figure*}[t]
    \centering
    \resizebox{0.85\linewidth}{!}{
    \includegraphics{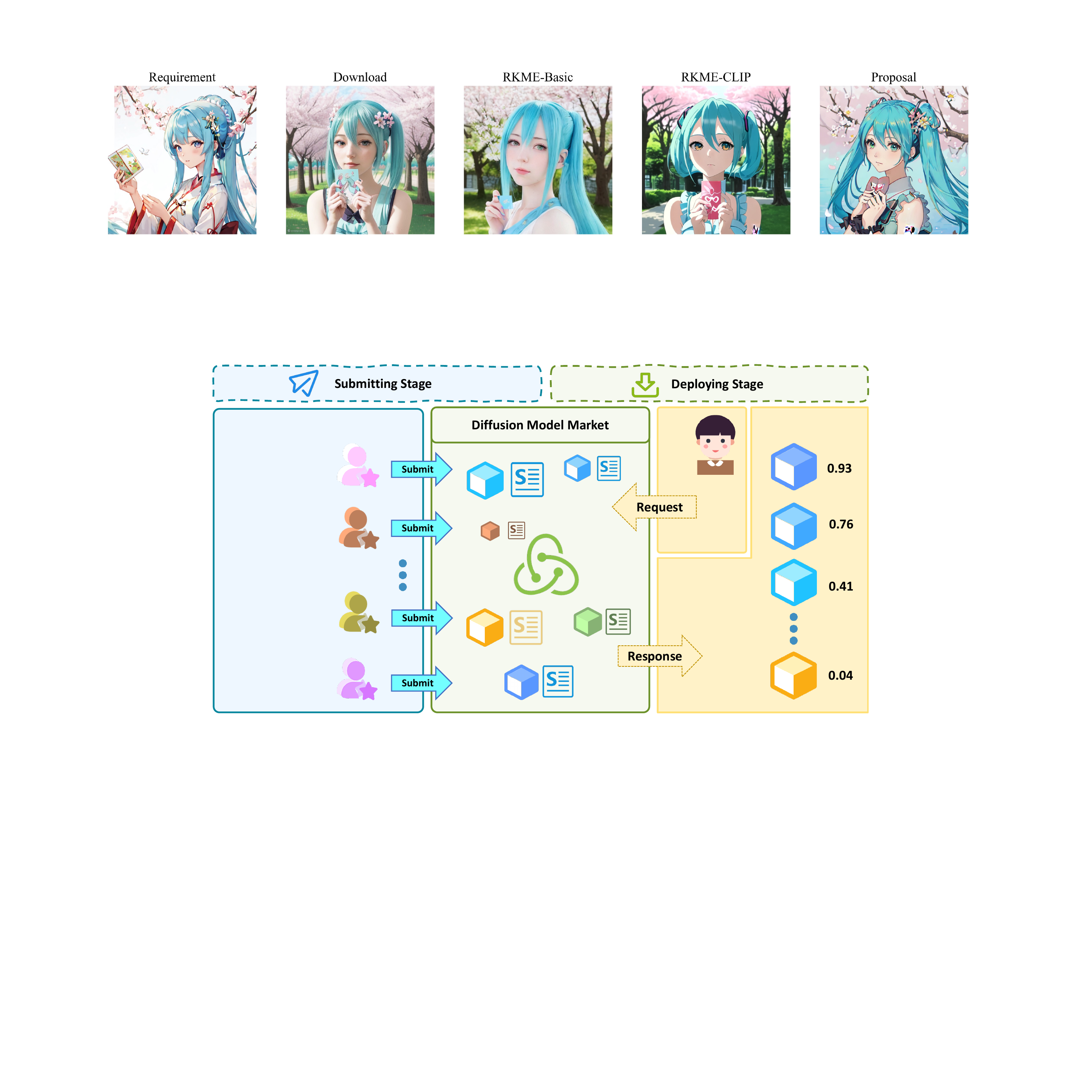}
    }
    \caption{Our proposed \algo method includes three main components: \emph{Automatic Specification Assignment}, \emph{Requirement Generation}, and \emph{Task-Specific Matching}. 
    \emph{Automatic Specification Assignment} generates a specification for each model using either a predefined or developer-provided prompt set to describe model functionality within the model matching space. \emph{Requirement Generation} formulates the requirement for each user task by encoding example images and their textual descriptions into the same space. \emph{Task-Specific Matching} adjusts the specification in the matching space according to the requirement and identifies the most suitable model with the highest similarity score.}
    \label{fig:method}
    \vspace{-0.3cm}
\end{figure*}

\section{Our Approach}

In this section, we present our solution \emph{\underline{P}rompt-Based \underline{M}odel \underline{I}ndetification} (\algo) for the \Setup\ setting. 
As illustrated in Figure~\ref{fig:method}, \algo consists of three key modules. \emph{Automatic Specification Assignment} and \emph{Requirement Generation} project the model's functionality and user requirements into a unified model matching space, addressing the first challenge. \emph{Task-Specific Matching} adjusts the specification in the matching space according to the requirement and identifies the most suitable model with the highest similarity score, addressing the second challenge.
We first describe the three key modules in detail. Then, a further analysis is provided as follows.

\subsection{Automatic Specification Assignment}

Automatic specification assignment aims to automatically generate a specification for each conditional generative model to describe its functionality. 
Its core idea is to prompt the conditional generative model $f_i$ to generate data for describing the model functionality based on a pre-defined prompt set $\PP$ or developer-provided prompt set $\PP_i$.

When model developer submit a model $f_i$ to the model hub, the automatic specification assignment algorithm $\cA_s$ generates its corresponding specification $S_i$ using a $N$-size prompt set $\PP$ which is pre-defined in the model hub. A set of images $X_i$ is generated by the model $f_i$ and the prompt set $\PP$ to describe model functionality conditioned on $\PP$: 
\begin{equation}
X_i = \left \{x^i_j = f(p_j) | p_j \in \PP \right \}_{j=1}^{N}.
\label{eq:spec-build-images}
\end{equation}
Then, the prompt set $\PP$ and the generated images $X_i$ are encoded to a unified model matching space using textual encoder $\cT(\cdot)$ and vision encoder $\cG(\cdot)$ from a pre-trained vision-language model respectively:
\begin{equation}
\begin{cases}
Z_i = \{ z^i_j = \cG(x^i_j)| x^i_j \in X_i\}, \\
Q_i = \{ q^i_j = \cT(p_j)| p_j \in \PP\}. \\
\end{cases}
\label{eq:spec-build-spec}
\end{equation}
Finally, the specification $S_i$ is defined as
\begin{equation}
S_i = \cA_s(f_i, \PP) = \{Z_i, Q_i\}.
\end{equation}

Note that if developer can provide a prompt set $\PP_i$ for model $f_i$ to better describe the model functionality, the better specification $S_i$ can be computed by replacing $\PP$ with $\PP_i$ in \autoref{eq:spec-build-images} and \autoref{eq:spec-build-spec}.

The advantages of our proposed automatic specification assignment are two-fold: (1) The specification $S_i$ can be automatically computed within the model hub, providing great convenience for developers and reducing their burden of uploading models. (2) The specification does not require a significant amount of storage space on the model hub, as it only involves storing the feature representation. 

\subsection{Requirement Generation}

Requirement generation aims to produce the requirements $R_{\tau}$ for the user task $\tau$ to select the most suitable model. 
Its core idea is to decompose the complex model functionality into the difference between user-provided example images $X_\tau=\{x^{\tau}_i\}_{i=1}^{N_\tau}$ and corresponding textual descriptions, enabling the future model identification to be more accurate.

Specifically, requirement generation algorithm $\cA_r$ transforms $X_{\tau}$ into feature representations:
\begin{equation}
    Z_{\tau}=\{ z^{\tau}_i = \cG(x^{\tau}_i)\}_{i=1}^{N_\tau}
\end{equation}
using the vision encoder $\cG(\cdot)$. 
Then, the textual description $\widehat{\PP}_{\tau}$ is generated by a vision-language model $\mathrm{VLM}(\cdot)$ to describe each example image and be mapped to the model matching space:
\begin{equation}
    \begin{cases}
    \widehat{\PP}_{\tau}=\{\hat{p}^{\tau}_i = \mathrm{VLM}(x^{\tau}_i)\}_{i=1}^{N_\tau} \\
    \widehat{Q}_{\tau} = \{\hat{q}^{\tau}_i = \cT(\hat{p}^{\tau}_i)\}_{i=1}^{N_\tau}
    \end{cases}
\end{equation}
Finally, the user requirement $R_{\tau}$ is computed using $\cA_r$ as
\begin{equation}
    R_{\tau} = \cA_r(X_\tau) = \left \{Z_{\tau}; \widehat{Q}_{\tau} \right \}.
\end{equation}

Note that $R_{\tau}$ is also automatically computed within the model hub, which is very flexible and easy to use.

\subsection{Task-Specific Matching}

Task-specific matching aims to identify the most suitable model for the user task $\tau$ by calculating the similarity score between the user requirement $R_{\tau}$ and the specification $S_m$ of each model $f_m$. 
Its core idea is to transform the specification $S_m$ corresponding to the user requirement $R_{\tau}$ in the model matching space to better match the functionality.

Specifically, matching algorithm $\cA_e$ calculates the similarity score between the user requirement $R_{\tau}=\{Z_{\tau}, \widehat{Q}_{\tau}\}$ and the specification $S_m=\{Z_m, Q_m\}$ for each model $f_m$ using the following formula:
\begin{equation}
\resizebox{1.0\linewidth}{!}{$\cA_{e}(S_m, R_{\tau}) = \frac{1}{N_\tau} \sum\limits_{i=1}^{N_\tau} \left \| \frac{1}{N_m} \sum\limits_{j=1}^{N_m} \frac{q^m_j \hat{q}^{\tau}_i}{\|q^m_j\| \|\hat{q}^{\tau}_i\|} k(z^m_j, \cdot) - k(z^{\tau}_i, \cdot) \right \|_{\cH_k}^2$}
\label{eq:score-proposal}
\end{equation}
where $\frac{q^m_j \hat{q}^{\tau}_i}{\|q^m_j\| \|\hat{q}^{\tau}_i\|}$ measures the similarity between the $j$-th specification prompts for model $f_m$ and the textual descriptions of example image $x^{\tau}_i$, transforming organial specifications to task-specific specifications, which are more focused on the user tasks. 
Then, the matching between task-specific specifications and user requirement can more accurately identify the most suitable model which meets the required model functionality.
Finally, the similarity score obtained by \autoref{eq:score-proposal} can be used to sort the models or directly return the most suitable model.

\subsection{Discussion}

It is evident that our proposal for the \Setup\ scenario achieves a higher level of accuracy and efficiency when compared to model search techniques employed by existing model hubs.

\paragraph{Accuracy.} Our proposal elucidates the functionalities of generated models by capturing both the distribution of generated images and prompts. This approach allows for more accurate identification of suitable models for users, as opposed to the traditional model search method that relies on download counts and star ratings for ranking models.

\paragraph{Efficiency.} Suppose that the model hub generates one requirement in $T_r$ time and calculates the similarity score for each model in $T_s$ time. The time complexity of our proposal for one identification is $O(T_r + M T_s)$ time. Moreover, with accurate identification results, users can save the efforts of browsing and selecting models, as well as reducing the consumption of network and computing. This is linearly correlated to the number of models on the model hub (which can be reduced by filtering by tags). Additionally, our approach also has the potential to achieve further acceleration through the use of a vector database~\cite{LW-Guo23} such as Faiss~\cite{Faiss-johnson2019billion}.

\begin{table*}[t]
    \begin{center}
    \resizebox{\linewidth}{!}{
    \begin{tabular}{l|c@{\hspace{0.3em}}c@{\hspace{0.3em}}c@{\hspace{0.3em}}c@{\hspace{0.3em}}c|c|c}
    \toprule
    Methods & Acc.($\uparrow$) & {\small Top-2} Acc.($\uparrow$) & {\small Top-3} Acc.($\uparrow$) & {\small Top-4} Acc.($\uparrow$) & {\small Top-5} Acc.($\uparrow$) & {\small Avg.} Rank$(\downarrow)$ & FID Score$(\downarrow)$ \\
    \midrule
    Baseline & ~~1.5\% & ~~3.0\% & ~~4.6\% & ~~6.1\% & ~~7.6\%  & 33.000 & 23.44 \\
    RKME & ~~3.1\% & ~~4.6\% & ~~6.2\% & ~~7.7\% & ~~9.3\%  & 32.014 & 25.47 \\ \hline
    \algo & \textbf{69.2\%} & \textbf{78.1\%} & \textbf{82.8\%} & \textbf{85.8\%} & \textbf{88.0\%} & ~~\textbf{2.874} & \textbf{18.42} \\
    \bottomrule
    \end{tabular}}
    \caption{Model Identification Performance of each method evaluated by Top-k accuracy, average rank and the generation quality evaluted by FID score when only one example image is provided. The results show that our \algo can achieve satisfactory model identification performance as well as generation quality. The best performance is in \textbf{bold}.}
    \label{tab:single}
    \end{center}
    \vspace{-1em}
\end{table*}

\begin{table}[t]
    \begin{center}
    \resizebox{\linewidth}{!}{
    \begin{tabular}{l|c@{\hspace{0.3em}}c@{\hspace{0.3em}}c|c@{\hspace{0.3em}}c@{\hspace{0.3em}}c}
    \toprule
       & \multicolumn{3}{c|}{Accuracy($\uparrow$)} & \multicolumn{3}{c}{FID Score($\downarrow$)} \\ 
       &  Baseline & RKME  & \algo     &  Baseline & RKME  & \algo \\
    \midrule
       1 image  &        1.5\% &         3.1\% & \textbf{69.2\%} & 23.44 & 25.47 & \textbf{18.42} \\
       2 images &        1.5\% &         3.1\% & \textbf{84.4\%} & 23.44 & 25.53 & \textbf{18.17} \\
       3 images &        1.5\% &         3.1\% & \textbf{89.7\%} & 23.44 & 25.53 & \textbf{18.14} \\
       4 images &        1.5\% &         3.2\% & \textbf{92.7\%} & 23.44 & 25.58 & \textbf{18.21} \\
       5 images &        1.5\% &         3.2\% & \textbf{94.0\%} & 23.44 & 25.61 & \textbf{18.18} \\
       6 images &        1.5\% &         3.2\% & \textbf{95.9\%} & 23.44 & 25.53 & \textbf{18.12} \\
    \bottomrule
    \end{tabular}}
\end{center}
\caption{Comparison of accuracy and FID score across methods with varying numbers of example images. The results demonstrate that \algo performance improves with additional examples, while RKME shows minimal change. The best performance is in \textbf{bold}.}
\label{tab:multi}
\vspace{-1em}
\end{table}

\section{Experiments}

To verify the effectiveness of our proposed method \algo for \Setup problem, we first build a novel conditional generative model identification benchmark based on stable diffusion models~\cite{SD-Rombach22}, and then conduct experiments on this benchmark. 
Below, we first introduce the details of the benchmark and evaluation metrics. 
Then, we present the experimental results on this benchmark and human evaluation results to emphasize the importance of \Setup problem as well as the effectiveness of our \algo.

\subsection{\Setup Benchmark}

In this section, we describe the our constructed \Setup\ benchmark and corresponding evaluation metrics. 

\paragraph{Model Hub and Task Construction.}

In practice, we expect model developers to submit their models to the model hub, allowing users to identify models that meet their specific needs. To enhance the realism of the evaluation, we constructed a model hub and user identification tasks to simulate this scenario. 
For the model hub construction, we manually collected $M=65$ different stable diffusion models $\{ f_1, \ldots, f_{m}, \ldots, f_{M} \}$ from CivitAI, representing uploaded conditional generative models on the hub. These models belong to the same category to mimic the real process where users first apply category filters before selecting models. 
For model specification generation, we created 61 prompts $\{ p_1, \ldots, p_{61} \}$ as a pre-defined prompt set $\PP$ in the model hub to simulate cases where developers do not provide specialized prompts. Additionally, we constructed 61 prompts $\{ p_{1}, \ldots, p_{61} \}_{m}$ for each model $f_m$ based on the prompts provided by developers in CivitAI to simulate developer-provided prompts. 
For model identification task construction, we created 14 evaluation prompts $\{ p_{\tau_1}, \ldots, p_{\tau_{14}} \}_{m}$ for each model on the model hub to generate testing images with random seeds in $\{0, 1, 2, 3, 4, 5, 6, 7, 8, 9\}$, forming $M_{\tau} = 14 \times 65 \times 10 = 9100$ different identification tasks $\left \{ (x_{\tau_i}, t_i) \right \}_{i=1}^{M_{\tau}}$, where each example image $x_{\tau_i}$ is generated by model $f_{t_i}$ and its best matching model index is $t_i$. 
We also provide ground-truth prompts for generating example images to assess the identification performance of each method in terms of the quality of generated images. We ensure that there is no overlap between prompts used for constructing specifications, guaranteeing the correctness of the evaluation.

\paragraph{Evaluation Metrics.}
In our experiments, we use Top-k accuracy, average rank, and FID score to evaluate the performance of methods.
We define the rank of model $f_m$ for task $\tau$ as $\widehat{r}^{\tau}_m = 1 + \sum_{i=1}^M \bI \left [ \widehat{s}^{\tau}_i < \widehat{s}^{\tau}_m \right ]$.
Then, the Top-k accuracy is defined as $\frac{1}{N^{\tau}} \bI \left [ \widehat{r}^{\tau_i}_{t_i} \leq k \right ]$, which evaluates the ability of each method to find the best matching model within $k$ trials.
The average rank is defined as $\frac{\widehat{r}^{\tau}_{t_i}}{N^{\tau}} $, which evaluates the ability of each method to rank the best matching model among other models.
Moreover, we use the identified model to generate images using the ground-truth prompt for generating example images for each task.
The FID score measures the distance between distribution of $M_{\tau}$ query images and distribution of $M_{\tau}$ generated images, evaluating the identification performance in aspects of generation quality. 

\subsection{Experimental Settings}

In this section, we introduce the experimental settings, including comparison methods and implementation details.

\paragraph{Comparison Methods.}
First, we compare our proposal with baseline method, which always uses the model with the highest downloading volume as the best-matched model~\cite{GPT-abs-2303-17580}. 
The performance of baseline can be identified by a reasonable lower bound of \Setup problem. 
Then, we also consider the basic implementation of the RKME specification~\cite{LW-Wu23} as a comparison method, namely, RKME, for the \Setup\ problem to evaluate whether learware techniques can be applied to the \Setup. 

\paragraph{Implementation Details}
We adopt the official code in~\cite{LW-Wu23} to implement the RKME method and the official code in~\cite{CLIP-RadfordKHRGASAM21} to implement the pre-trained vision-language model.
We follow the default hyperparameter setting of RKME in previous studies~\cite{LW-Guo23}, setting the size of the reduced set to 1 and choosing the RBF kernel~\cite{RBF-XuKY94} for RKHS. The hyperparameter $\gamma$ for calculating RBF kernel and similarity score is tuned from $\left \{0.005, 0.006, 0.007, 0.008, 0.009, 0.01, 0.02, 0.03, 0.04, 0.05\right \}$ and set to $0.02$. 
For all experiments without additional notes, we assume that the specification is generated with developer-provided prompts. 
Our experiments are conducted on Linux servers with NVIDIA A800 GPUs. 

\subsection{Experimental Results}

In this section, we present the experimental results of our \algo method and comparison methods. 

\paragraph{Model Identification Performance.}
We evaluate the model identification performance of each method in Table~\ref{tab:single} when only one example image is provided. 
The Top-k accuracy and average rank metrics quantify how well each method identifies the optimal model. 
The FID score measures the quality of images generated by the identified models using ground-truth prompts.
Results show that RKME performs similarly to the baseline method, with poor accuracy and rank, indicating the inherent difficulty of the \Setup problem and the limitations of existing techniques designed for discriminative models for \Setup problem. 
Our \algo significantly outperforms RKME in both accuracy and rank metrics, leading to two key findings: 
(1) the \Setup problem can be effectively solved with appropriate model identification strategies; 
(2) \algo provides satisfactory performance even with one single example image.
The FID scores further demonstrate that models identified by \algo generate higher quality images, confirming that one fixed popular model cannot meet all use cases and highlighting the significance of the \Setup problem.
Table~\ref{tab:multi} presents results with multiple example images.
The performance of \algo improves with additional examples, while RKME's performance remains largely unchanged, validating the effectiveness of our approach.

\begin{table}
    \begin{center}
    \begin{tabular}{cc|ccc}
    \toprule
      MMS & TSM &  FID $(\downarrow)$ & Accuracy $(\uparrow)$ & Rank $(\downarrow)$ \\
    \midrule
      & & 25.47 & ~~3.1\% & 32.014 \\
      \checkmark & & 18.43   & 68.0\% & ~~3.120 \\
      \checkmark & \checkmark & \textbf{18.42} & \textbf{69.2\%} & ~~\textbf{2.874} \\
    \bottomrule
    \end{tabular}
    \caption{Ablation study. MMS indicates the model matching space, which includes the automatic specification assignment and the requirement generation. TSM indicates the task-specific matching. The best performance is in \textbf{bold}.}
    \label{tab:ablation}
\end{center}
\vspace{-1em}
\end{table}

\paragraph{Ablation Study.} 
To analyze the contribution of each component in \algo, we conduct an ablation study as shown in Table~\ref{tab:ablation}. 
Our \algo method extends the RKME framework with two groups of core components:
(1) Model Matching Space (MMS), which consists of the automatic specification assignment algorithm $\cA_s$ and the requirement generation algorithm $\cA_r$. These algorithms project both user requirements and model functionality into a unified matching space.
(2) Task-Specific Matching (TSM), which introduces algorithm $\cA_t$ to perform precise model functionality matching.
The results demonstrate that integrating both components is essential for achieving optimal performance.

\paragraph{Human Evaluation and GPT-4o Evaluation.}

\begin{figure}
    \centering
    \includegraphics[width=\linewidth]{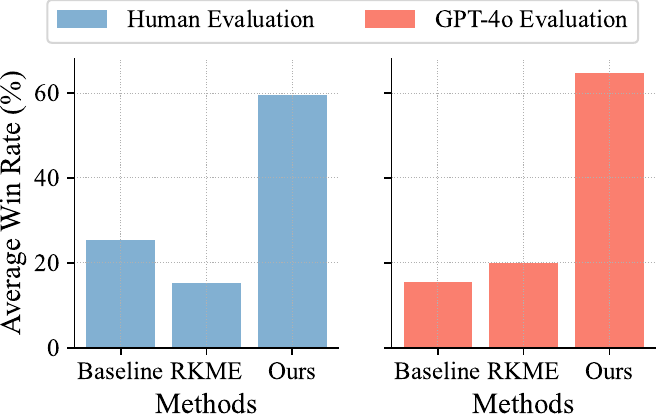}
    \caption{Human evaluation results. The results show that human users significantly prefer the images generated by our proposal.}
    \label{fig:human_eval}
\end{figure}

To further validate the effectiveness of our \algo, we conducted human evaluation with 70 users and a GPT-4o evaluation, respectively
For human evaluation, each user completed a survey containing 5 questions and each question is randomly sampled from 9100 tasks (we skip tasks where all methods identify the same model) using images generated by Baseline, RKME, \algo methods as the options. 
The users are required to select the image that best matches the example image. For GPT-4o evaluation, we prompt GPT-4o to choose the best image from the options for all 9100 tasks. 
The \autoref{fig:human_eval} presents the average win rate of each method voted by human users and GPT-4o, respectively. 
The results show that human users and GPT-4o have different preferences for the images generated by the Baseline method and RKME method, giving different yet similar win rates for these two methods. 
Our \algo achieves the highest win rate with a large margin, indicating that the images generated by our \algo are more consistent with the user requirements.

\subsection{Further Analysis}

In this section, we analyze the model identification performance when using default prompts and present qualitative results through visualization.

\paragraph{Default Prompt Set.} 
When model developers do not provide prompts, the model hub uses a default prompt set to generate specifications, making it more adaptable to different conditional generative models. 
Table~\ref{tab:default-prompt-fid} shows the FID scores when using default prompts for all models. 
Our \algo achieves significantly lower FID scores compared to RKME and baseline methods, demonstrating its ability to generate high-quality images even with pre-defined prompts.
Table~\ref{tab:default-prompt-acc} presents the Top-k accuracy with one example image. 
While \algo maintains superior performance over RKME, its accuracy is lower compared to using developer-provided prompts in Table~\ref{tab:single}, indicating that developer-provided prompts are valuable for optimal model identification.

\begin{table}[t]
    \begin{center}
    % \resizebox{\columnwidth}{!}{
    \begin{tabular}{l|rrrr}
    \toprule
    Methods &  1 image &  2 images &  3 images &  4 images \\
    \midrule
      Baseline &  23.44 &  23.44 &  23.44 &  23.44 \\
      RKME &  38.24 &  38.31 &  38.31 &  38.34 \\ \hline
      \algo &  \textbf{20.05} &  \textbf{19.94} &  \textbf{19.94} &  \textbf{20.13} \\
    \bottomrule
    \end{tabular}
    \end{center}
    \caption{FID score of each method when all models use default prompts set to generate specifications when different number of images are provided. The best performance is in \textbf{bold}.}
    \label{tab:default-prompt-fid}
    \vspace{-1em}
\end{table}

\begin{table}[t]
\begin{center}
\resizebox{\linewidth}{!}{
\begin{tabular}{l|cccc}
\toprule
Methods &  Acc. ($\uparrow$) &  {\small Top-2} Acc. ($\uparrow$) &  {\small Top-3} Acc. ($\uparrow$) &  {\small Top-4} Acc. ($\uparrow$) \\
\midrule
Baseline &  ~~1.5\% &  ~~3.1\% &  ~~4.6\% &  ~~6.2\% \\
RKME    &  ~~3.1\% &  ~~4.6\% &  ~~6.2\% &  ~~7.7\% \\
\hline
\algo &  \textbf{27.9\%} & \textbf{38.5\%}  &  \textbf{45.3\%} &  \textbf{50.8\%} \\
\bottomrule
\end{tabular}}
\end{center}
\caption{Top-k accuracy of Baseline, RKME, and \algo methods when all models use default prompts set to generate specifications. The best performance is in \textbf{bold}.}
\label{tab:default-prompt-acc}
\end{table}

\begin{figure}[t]
    \begin{center}
    \centerline{\includegraphics[width=\linewidth]{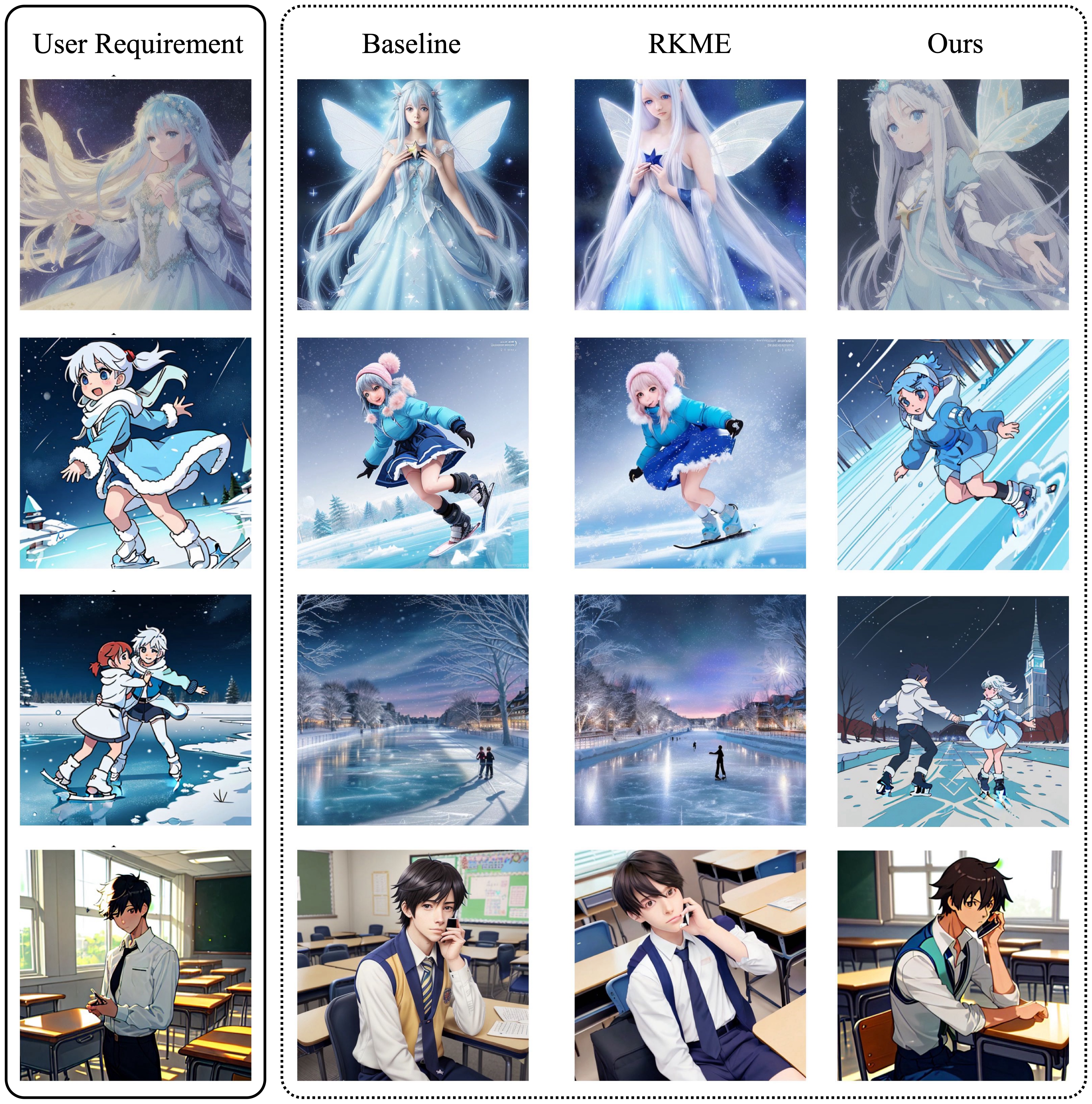}}
    \caption{Visualization of generated images from models identified by Baseline, RKME, and \algo. The results demonstrate that \algo identifies models that can generate images with styles consistent with the example images.}
    \label{fig:examples}
    \end{center}
    \vspace{-2em}
\end{figure}

\paragraph{Visualization.}
We visualize the generated images from models identified by Baseline, RKME, and \algo in Figure~\ref{fig:examples}, with example images shown in the first column. 
While all identified models generate images with correct content, they differ significantly in style. 
Our \algo successfully identifies models that match the comic art style of the example images, whereas models identified by other methods generate images that are overly realistic.
These results show that our \algo is helpful for generating images similar to the example images, which is consistent with the experimental results. 

\section{Conclusion}

In this paper, we study a novel problem setting called \emph{Conditional Generative Model Identification}, whose objective is to describe the functionalities of conditional generative models and enable the model to be accurately and efficiently identified for future users. 
To this end, we present a systematic solution including three key components. The \emph{Automatic Specification Assignment} and \emph{Requirement Generation} respectively project the model functionality and user requirements into a unified matching space. 
The \emph{Task-Specific Matching} further builds the task-specific specification in the matching space to precisely identify the most suitable model. 
To promote relevant research, we open-sourced a benchmark based on stable diffusion models with 65 conditional generative models and 9100 model identification tasks. Extensive experiment results on the benchmark as well as the human evaluation demonstrate the important value of the \Setup\ problem and the effectiveness of our proposal. 

In future work, we intend to develop a novel generative model hub using the techniques presented in this paper. Our goal is to offer a more accurate description of conditional generative model functionalities and user requirements. We expect that this will enhance the efficiency of users in finding models that meet their specific needs and contribute to the development and widespread use of generative models, as well as promote the development of model hubs. 

One limitation of our work is that we only consider the cases that identify generative models using uploaded example images to describe users' requirements. The assumption is reasonable since users' ideas often rely on existing image templates when they want to generate images, and it is not difficult to find images that have a similar style to fulfill the user's requirements. Despite this, it is also interesting to study how to quickly and accurately identify models via other information such as textual prompts. 

\section*{Acknowledgements}
This research was supported by the National Natural Science Foundation of China (Grant No. 624B2068, 62306133, 62250069), the Key Program of Jiangsu Science Foundation (BK20243012), the Jiangsu Science Foundation (BG2024036), and the Fundamental Research Funds for the Central Universities (022114380023).

\section*{Contribution Statement}

Zhi Zhou and Hao-Zhe Tan contributed equally to this work.

\bibliographystyle{named}
\bibliography{ref}

\end{document}